\documentclass[conference]{IEEEtran}
\IEEEoverridecommandlockouts
\usepackage{cite}
\usepackage{amsmath,amssymb,amsfonts}
\usepackage{algorithm,algpseudocode}
\usepackage{graphicx}
\usepackage{textcomp}
\usepackage{xcolor}
  \usepackage{caption}
  \usepackage{comment}
  \usepackage{lipsum}
\usepackage{booktabs}
\usepackage{subcaption}
\def\BibTeX{{\rm B\kern-.05em{\sc i\kern-.025em b}\kern-.08em
    T\kern-.1667em\lower.7ex\hbox{E}\kern-.125emX}}
\begin{document}

\title{Massive Dimensions Reduction and Hybridization with Meta-heuristics in Deep Learning}

\author{\IEEEauthorblockN{}}
\author{
\IEEEauthorblockN{Rasa Khosrowshahli$^{1}$ Shahryar Rahnamayan$^{2}$, SMIEEE, Beatrice Ombuki-Berman$^{3}$}\\
\IEEEauthorblockA{
\textit{$^{1, 2}$Nature-Inspired Computational Intelligence (NICI) Lab}\\
\textit{$^{1, 3}$Bio-Inspired Computational Intelligence Research Group (BICIG)}\\
\textit{$^{1, 3}$Department of Computer Science, Brock University, St. Catharines, ON, Canada}\\
\textit{$^2$Department of Engineering, Brock University, St. Catharines, ON, Canada}\\
$^1$rkhosrowshahli@brocku.ca, $^2$srahnamayan@brocku.ca, $^3$bombuki@brocku.ca\\
}
}

\maketitle

\begin{abstract}
Deep learning is mainly based on utilizing gradient-based optimization for training Deep Neural Network (DNN) models.
Although robust and widely used, gradient-based optimization algorithms are prone to getting stuck in local minima. 
In this modern deep learning era, the state-of-the-art DNN models have millions and billions of parameters, including weights and biases, making them huge-scale optimization problems in terms of search space. Tuning a huge number of parameters is a challenging task that causes vanishing/exploding gradients and overfitting; likewise, utilized loss functions do not exactly represent our targeted performance metrics. 
A practical solution to exploring large and complex solution space is meta-heuristic algorithms. 
Since DNNs exceed thousands and millions of parameters, even robust meta-heuristic algorithms, such as Differential Evolution, struggle to efficiently explore and converge in such huge-dimensional search spaces, leading to very slow convergence and high memory demand. 
To tackle the mentioned curse of dimensionality, the concept of blocking was recently proposed as a technique that reduces the search space dimensions by grouping them into blocks. 
In this study, we aim to introduce Histogram-based Blocking Differential Evolution (HBDE), a novel approach that hybridizes gradient-based and gradient-free algorithms to optimize parameters.
Experimental results demonstrated that the HBDE could reduce the parameters in the ResNet-18 model from 11M to 3K during the training/optimizing phase by meta-heuristics, namely, the proposed HBDE, which outperforms baseline gradient-based and parent gradient-free DE algorithms evaluated on CIFAR-10 and CIFAR-100 datasets showcasing its effectiveness with reduced computational demands for the very first time. 
\end{abstract}


\section{Introduction}

In recent years, researchers have invested their time and efforts in proposing variants of deep neural networks (DNNs) to solve real-world problems effectively. Training DNNs is commonly done using gradient-based optimizers such as Adam \cite{kingma2014adam, gupta2023explainable}. 
Gradient-based optimization is a conventional method in training neural networks that necessitates the utilization of differentiable activation and loss functions. This approach involves the computation of the gradient of the objective function with respect to the model parameters, which is subsequently employed to update such parameters.

There is a strong relation between the choice of differentiable loss function and overfitting and a weak relation between the differentiable loss function and evaluation metric. In this way, the model becomes more prone to outliers in the training set if the objective function penalizes significant differences between the predicted and target labels because of a huge gap between data distributions. However, the ultimate objective is to achieve the expected performance on evaluation metrics such as F1-score on the unseen (test) data.


An alternative approach is using gradient-free optimization algorithms. These methods do not require calculating gradients and are helpful when computing gradients is impractical and complex. 
Thus, the objective function could be any, such as an evaluation metric (e.g., F1-score, Precision, Recall, etc.). 
These methods explore the behavior of objective function by iteratively evaluating different points in the parameter space and adjusting the search direction based on the function values observed. One of the well-known families of gradient-free optimization algorithms is meta-heuristic algorithms, designed to be versatile and can be applied to various optimization problems. 
In the literature, meta-heuristic algorithms are applied to optimize DNNs for different purposes, such as weights and biases, the number of layers and neurons in the layers, and hyper-parameter settings. 
Based on a recent comprehensive review by Kaveh \textit{et al.}\cite{kaveh2023application}, just $20\%$ of studies worked on optimizing small-scale weights and biases in various neural networks, showing an open research direction. Hybridization of gradient-based optimization and meta-heuristic algorithms has been demonstrated to boost accuracy \cite{al2023boosting, ansari2020hybrid, HybridPSOCNN, ganjefar2017training, yaghini2013hybrid}.
These algorithms are inspired by metaphors from nature, such as evolution, swarm intelligence, or annealing. 
To increase the accuracy of Convolutional Neural Networks, Rere \textit{et al.}\cite{rere2016metaheuristic} presented the techniques of Differential Evolution, Harmony Search, and Simulated Annealing algorithms, which fall under meta-heuristics \cite{akay2022comprehensive}. 
Recently, Rokhsatyazdi \textit{et al.}\cite{Rokhsatyazdi2023ANNCS} proposed a two-extreme-point Coordinate Search (CS) algorithm to optimize a relatively small fully connected network with two hidden layers consisting of 266K parameters evaluated on the MNIST handwritten digits dataset. 
They reported that the learning convergence rate of CS as a meta-heuristic algorithm is faster than the Stochastic Gradient Descent (SGD) algorithm for such a huge-dimensional optimization problem.
Due to the scarcity of dimensionality in the search space of neural network parameters, previous works are limited by the memory resources and cost to utilize gradient-free optimizers in training or fine-tuning neural networks. 
In the case of training SOTA deep neural network architectures such as ResNet \cite{He_2016_CVPR}, the number of trainable parameters is in the size of millions and even recently reached billions in large language models such as GPT-2 \cite{radford2019language}.
The key idea is to reduce the trainable parameters in the neural networks without dropping layers or compressing the architecture. 
In the literature, the weight-sharing technique discusses giving control of several connections between layers in neural networks by a single parameter as proposed by Rumelhart  \cite{rumelhart1986general}. 
In other words, the number of free parameters is reduced while keeping the size of networks.
Later, LeCun \textit{et al.} \cite{lecun1989generalization} generalized the idea into Weight Space Transformation (WST), in which the strength of connections is computed from a transformation of parameters, leading to improvement in learning speed and reducing the size of parameters space.
Nowlan and Hinton \cite{nowlan2018simplifying} proposed a clustering-based weight-sharing approach in which the closest value-wise weights are located in different clusters. The purpose of their study was to improve generalization performance than relying on weight decay or a learning time controller technique.

In this work, we aim to use gradient-free optimization to tune the weights/biases and evaluate them, but since the number of trainable parameters is huge, it is required to reduce the dimensionality of the search space. We introduce a novel idea to reduce the number of parameters in training by sharing the adjusted weights/biases into the blocks of parameters. To be exact, our contributions can be summarized as follows:
\begin{itemize}
    \item We review the meta-heuristic algorithm, Differential Evolution, an evolutionary algorithm used to optimize large-scale optimization problems. Since the population in evolutionary algorithms takes $O(NP \times D)$ memory space, we reviewed the Block DE algorithm, which significantly reduces both memory and search space.
    \item We introduce a hybridization of deep neural network training with gradient-based and gradient-free optimization algorithms to fine-tune the parameters.
    \item Through experiments on image classification, we show that block-ing trainable parameters leverage the convergence speed and efficacy of memory space in gradient-free optimization, especially evolutionary optimization. Nevertheless, avoiding gradients allows us to use evaluation metrics as an objective function, which decreases the gap between training and testing.
    \item We achieved better performance (reducing the cost of search) from $D=11M$ to $D=3.5K$, which we attribute to the use of gradient-free evolutionary optimization. 
\end{itemize}

The rest of the paper is organized into the following procedure.
Section \ref{sec:background} reviews the Differential Evolution meta-heuristic algorithm and the Block technique on Differential Evolution. Section \ref{sec:proposed} outlines the proposed dimension reduction and hybrid optimization, which consists of several components. Section \ref{sec:experiments} presents conducted experiments on fine-tuning ResNet-18 model by the proposed method on CIFAR-10 and CIFAR-100 datasets. We conclude the paper in Section \ref{sec:conclusion}.

\section{Backgroud Review}\label{sec:background}
\subsection{Differential Evolution}
One of the evolutionary population-based algorithms that gained popularity and drew attention from numerous optimization researchers is Differential Evolution (DE) \cite{DE1997}. The method should begin with a produced initial population, as with other evolutionary algorithms.  The DE uses the crossover and mutation operators to create new trial vectors. The mutation operator calculates a vector $\hat{y}_i$, or a mutant vector, using a linear combination function and randomly selected candidate solutions of the current population. The following is the linear combination function:
\begin{equation}
\label{eq:rand1}
\hat{y}_i = x_{i1} + F*(x_{i2} - x_{i3}).
\end{equation}
where ${x_{i1}}$, ${x_{i2}}$, and ${x_{i3}}$ are three randomly selected member from the population and $F$ is the scaling factor to control the differentiation. By exchanging the mutant vector $x_{i}$ with its corresponding parent (target) vector $x_i$, the crossover operator generates a new vector $y_i$, also referred to as the trial vector. When calculating $y_i$, one of the most well-known crossovers is binomial formulated as follows:
\begin{equation}
y_{i,j}= \left\{ \begin{array}{rcl} \hat{y}_{i,j} & rand_{i,j} \leq CR\ or\ j == j_{rand} \\ x_{i,j} & otherwise \end{array}\right.
\label{eq:bin_crossover}
\end{equation}
where $j=1,...,D$, an integer random number $j_{rand}$ in $1,...,D$, a real random number $rand_{i,j} \in [0,1]$, and the crossover probability $CR \in [0,1]$ are all present.
Finally, the selection operator compares the associated target vector $x_{i}$ with trial vector $v_{i}$ and, after computing fitness values, chooses the most fit among them to represent the people of the following generation. 

\subsection{Block Differential Evolution}
Recently, a novel optimization algorithm called Block Differential Evolution (BDE) has been proposed by Khosrowshahli and Rahnamayan \cite{KhosrowshahliBlockDE2023} that addresses spatial issues in huge-scale problems and aims to reduce the convergence time of the Differential Evolution (DE) algorithm. The block scheme is a one-time blind clustering of dimensions in a search space, which was motivated by Block Coordinate Descent (BCD) and Discrete Coordinate Descent (DCD) algorithms \cite{9283201}. 
The primary motivation behind BDE is to overcome challenges associated with the memory requirements of embedded systems through a huge dimension reduction. 

In conventional DE, the algorithm operates on the entire dimensionality of the problem, which can be computationally demanding for large-scale problems. BDE proposes a dimension reduction-based approach where specific grouped or blocked parts of the dimensions are saved and optimized using operators. The remaining dimensions are blocked from access, resulting in a more efficient optimization process. 

The key steps in their proposed scheme involve defining the problem's dimension as $D$ and the block size as $BS$. The new blocked dimension size $D^{'}$ is then calculated as $D^{'} = D/BS$. An initial population for DE is generated with $NP$ individuals and $D$ dimensions. A dimension reduction process is employed to reduce the number of decision variables in each individual, resulting in a new search space with smaller dimensions. A single-time clustering process achieves the dimension reduction to block (group) dimensions with equal sizes and replaces the first dimension in each block with the entire block inside the decision vector. 
The operators in the DE algorithm work with the reduced size of candidate solutions, which is cost-effective in terms of memory, thus resulting in faster exploration and exploitation in terms of performance. 
BDE generates a new blocked population with dimensions $NP \times D{'}$. However, a method called Unblocker is introduced for fitness evaluation. This method reconstructs the original dimension of the problem by copying each dimension of the solution $BS$ times into a solution vector, which is then evaluated by the problem's fitness function.

During the operational stage of optimization, the population consistently maintains the alternative size of $NP \times D{'}$, utilizing memory that is $BS$ times smaller than the original size for each individual. This results in a smaller memory size for each generation that is $NP \times (BS-1)$ times less than the classic DE algorithm. 
The proposed BDE algorithm is positioned as a simple yet effective solution for optimization in limited-memory embedded systems.

\section{Proposed Algorithm}\label{sec:proposed}
This section describes the elements of the proposed hybrid of gradient-based and gradient-free DNN optimization process.

\subsection{Motivation: Meta-heuristic algorithms for DNNs}

In the pursuit of optimizing Deep Neural Networks (DNNs), one can employ gradient-free optimization algorithms falling within the realm of meta-heuristic algorithms. The parameter space of DNNs comprises the weights and biases associated with the connections between neurons, and a high dimensionality typically characterizes it.
Meta-heuristic algorithms may struggle to explore and exploit such large search spaces efficiently, leading to slow convergence or getting stuck in local optima.
Meta-heuristic algorithms commonly depict the complete parameter set of a neural network as a singular vector. This vectorization simplifies the optimization task, adjusting all parameters uniformly and facilitating the use of conventional optimization methods.
One solution involves decreasing the weights and biases within the solution vector. This allows the application of meta-heuristic algorithms, such as evolutionary algorithms, which function on compact candidate solution populations, enhancing exploration and exploitation within extensive search spaces.

We claim that gradient-free optimization has many advantages in training deep neural networks. 
Two elements need to be defined for the gradient-free optimization procedure to optimize a DNN's parameters effectively: the search space and the objective function.

\textbf{Search Space.}
A DNN architecture includes various layers, such as the convolutional layers and fully connected layers, which collectively enable the network to learn intricate patterns and representations from input data during the training process. 
The connections between each layer have trainable parameters called weight and bias, which are adjusted during the training process through optimization algorithms like gradient descent. These parameters play a crucial role in fine-tuning the model's ability to make accurate predictions by adjusting the strength and direction of the connections between neurons.
In Gradient-Free Optimization (GFO), the set of all weights and biases is organized into a 1-$D$ vector, commonly referred to as the parameter vector or parameter set. In this search space, the $D$ is defined as the size of the dimensions.
The GFO algorithms work directly with this parameter vector to explore and search the high-dimensional space of possible configurations. The optimization algorithm does not rely on gradient information, which is the key distinction from gradient-based optimization methods. Instead, it explores the parameter space based on the evaluations of the objective function at various points, adapting its search strategy to improve the model's performance iteratively. It is worth noting that using a 1-$D$ parameter vector is a convenient representation. GFO algorithms treat the neural network parameters as a flat, continuous vector, and the optimization process is guided by the observed performance of different configurations in this parameter space.

\textbf{Objective Function.} 
In conventional deep learning, a derivable function calculates the loss of DNNs on a given task, and a gradient-based optimization algorithm is used to update parameters on backward propagation during training.
However, the final performance is assessed by an evaluation metric to find the effectiveness or accuracy of the trained (DNN) model on the given task.
Since the evaluation metrics such as accuracy, precision, recall, and F1-score are non-differentiable functions, the constraints related to gradient computation are impossible to use in the gradient-based optimization approach.
This motivated us to choose a meta-heuristic algorithm to solve the optimization as it is robust to solve large-scale global optimization problems where the nonlinear and non-differentiable continuous space and the size are large and, therefore, considered widely used gradient-free methods. 
In order to handle the multi-class classification, we calculated the F1-score, which takes all classes to be equally important. 
The F1-score is calculated according to the following equation:
\begin{equation}
\label{eq:f1score}
    f(x) = \textit{F1-score} = \frac{2 \times Precision \times Recall}{Precision + Recall}.
\end{equation}
where $Precision$ can be calculated as follows

\begin{equation}
\label{eq:PR}
\mathit{Precision}= \frac{\text{True Positive}}{\text{True Positive} + \text{False Positive}}.
\end{equation}

\noindent Similarly, $Recall$  can be given as follows

\begin{equation}
\label{eq:RE}
\mathit{Recall}= \frac{\text{True Positive}}{\text{True Positive} + \text{False Negative}}.
\end{equation}
Ultimately, the F1-score metric is stated as the single-objective function $f(x)$ in our gradient-free algorithm.

\subsection{Training Stages by Gradient-free Optimization}
A common problem in training a deep neural network is facing overfitting when the model learns to perform well on the training data but fails to generalize effectively on unseen data. It is more expected to happen when the model becomes too complex, or the number of adjustable parameters is large relative to the size and complexity of the data. In this section, we adopt two approaches to reduce the overfitting and increase the convergence rate of meta-heuristic algorithms by utilizing pre-training and random super-batches as follows:

\textbf{Pre-training.} 
A simple step to accelerate the optimization is to pre-train the model. The parameters in a pre-trained model are fitted to the vast majority of the training dataset, but the model cannot predict the minority of data. For only the pre-training stage of a DNN model, we aim to use Adam, which is a gradient-based optimization algorithm. Adam \cite{kingma2014adam} is a well-known optimization algorithm in training DNN models, and it performs effectively for problems with a large number of data and/or parameters. As a matter of fact, Adam helps us accelerate the optimization to fit the most training data but not overfitting. However, there are some issues associated with using Adam, as follows:
1) Despite the fact that Adam has few parameters to be set, its sensitivity to the learning rate is deniable,
2) Utilizing an adaptive learning rate is good but causes a lack of orthogonality in model parameters, leading to poor generalization on unseen data.
and, 3) Adam might not perform on noisy or non-stationary loss functions resulting in premature convergence.
Because of this, pre-training model by Adam is halted after a few training epochs to avoid dropping on the overfitting problem. Gradient-free optimization is the next component that takes the position of the gradient-based optimizer, Adam, to fine-tune the model by exploring the search space. 

\textbf{Circular Super-batches.}
As a part of gradient-free optimization, we designed a data sampling system to replace super-batches randomly. In other words, the objective evaluation in each iteration is calculated based on randomly selected super-batch data for all the $NP$ population without overlapping with super-batches in the previous iterations. To this end, the candidate solutions in the population are fairly evaluated by cycles of the training dataset after iterations. 
Bodner \textit{et al.}\cite{bodner2021gradfreebits} argued that using moving (changing) super-batches could reduce the variance in the objective function, leading to an acceleration in the convergence rate of the meta-heuristic algorithm.

\subsection{Histogram-based Blocking Differential Evolution (HBDE)}

In this context, to reduce the search space complexity, we propose a parameter-reduction technique for deep neural networks inspired by the Block approach proposed by Khosrowshahli \textit{et al.}\cite{KhosrowshahliBlockDE2023}. Our proposed algorithm has two leading steps: pre-training the DNN model with a gradient-based optimizer and fine-tuning with a gradient-free optimizer. The pre-training allows the model to learn useful representations from a large dataset quickly. Once the pre-training is complete, the model is ready to transfer the knowledge gained during the pre-training stage to the target task by fine-tuning the model on the dataset using a meta-heuristic algorithm such as DE. 

\textbf{Histograms.} A histogram is commonly utilized in the process of data summarization, providing an approximate value, $\hat{p}(x)$ for the initial measurement of a variable $x$, as $p(x)$, particularly in the context of computing efficiency \cite{ng2005projective}.
The algorithm for creating a histogram of 1-$D$ dimensional vectors divides each dimension into a set of equitized $bins$ with bin width as $h$ typically involves the following formula:
\begin{equation}
    \hat{p}(x) = \frac{1}{N \times h}.
\end{equation}
Where $N$ is the total number of variables in the 1-$D$ vector. 
The bin width is a crucial histogram parameter since it manages the trade-off between under- and over-smoothing the actual distribution.
The histogram begins with calculating bin width, which determines the width of each bin by dividing the range of min and max weights, $l_W = min(W), u_W = max(W)$, by the number of bins $N_{bins}$ as follows:
\begin{equation}
    h = (u_W - l_W)/N_{bin}.
\end{equation}
Where each bin represents a specific range of $[l_{b_i}, u_{b_i}]$ for $bin_{i}$ which $i$ indicates the dimension. 
By having the lower and upper bounds for each $w_i$, uniform random distribution is used to initialize the candidate solutions population. We discard the empty bins and reduce the number of blocks from $N_{bins}$ to $N_{blocks}$, and in this way, the number of parameters in candidate solutions could be even less. In other words, the $N_{bins}$ represents the expected search space dimension for DE to explore; however, the real number of blocks $N_{blocks}$ optimally is smaller than $N_{bins}$.
\begin{figure}[]
    \centering
    \includegraphics[scale=1.0]{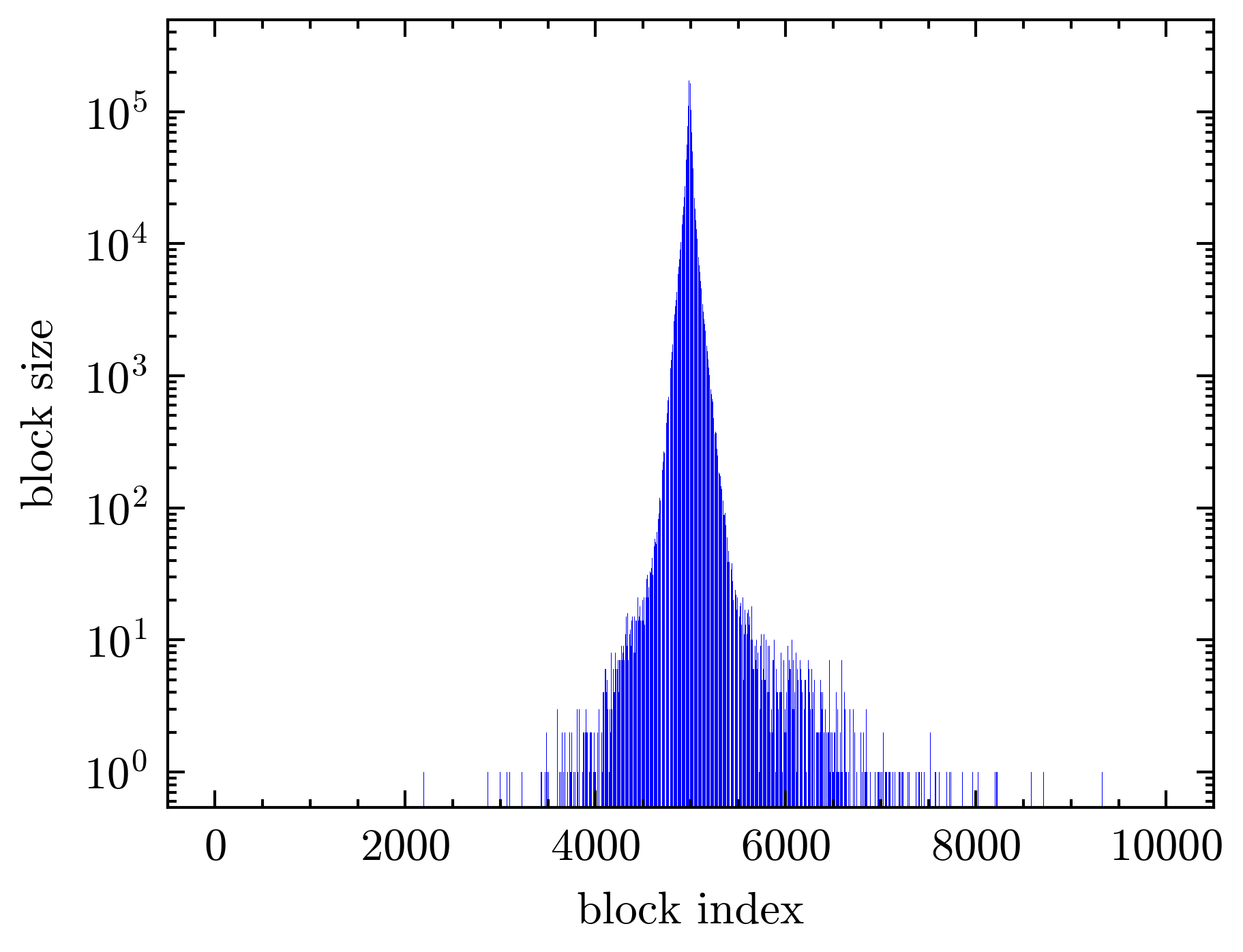}
    \caption{ResNet-18 parameters (11.2M) blocked by Histogram with $N_{max}=10,000$.}
    \label{fig:histogram_cifar10_10000B_before_remove_empty_bins}
\end{figure}

\begin{figure}[]
    \centering
    \includegraphics[scale=1.0]{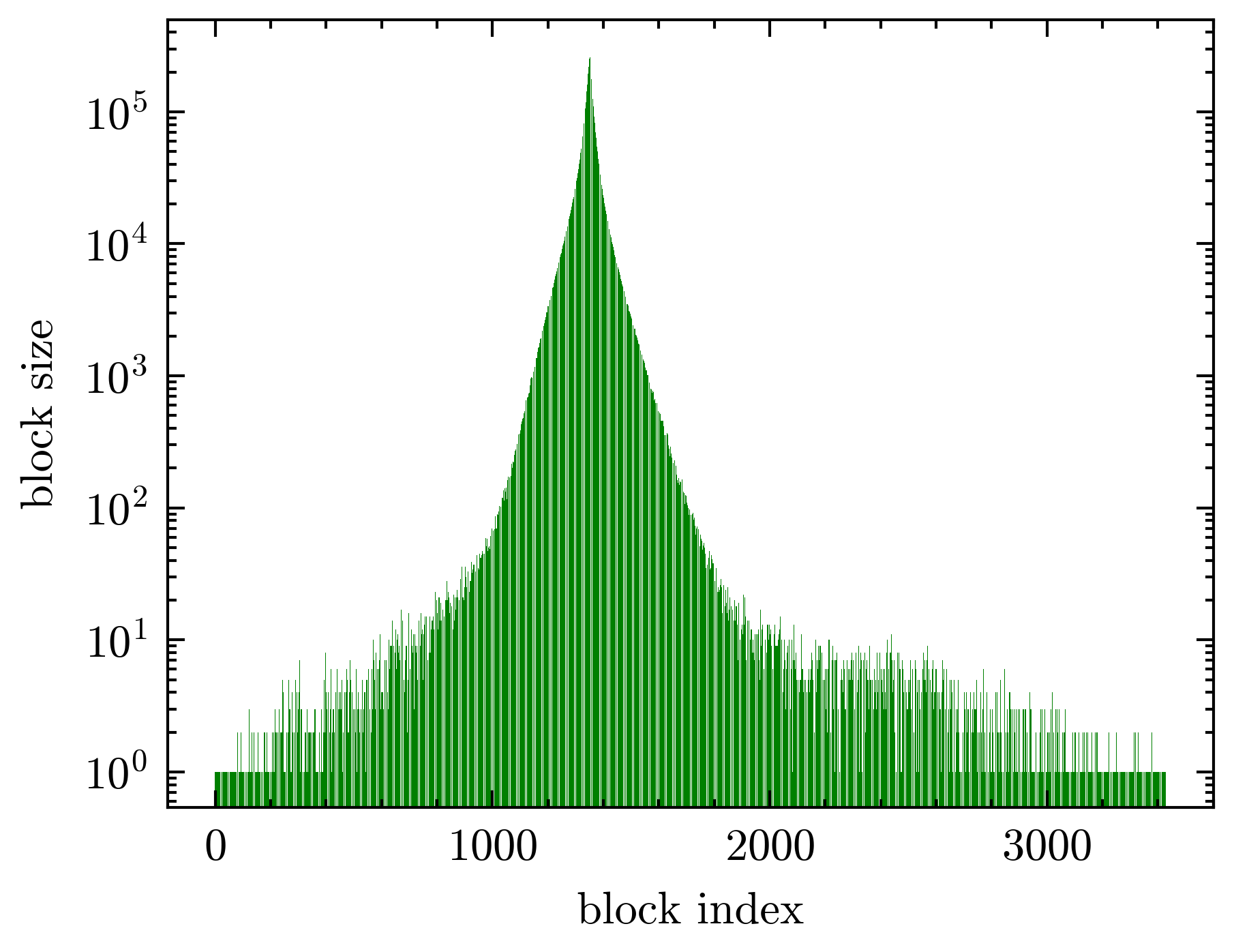}
    \caption{ResNet-18 parameters (11.2M) after removing empty blocks. The final number of blocks dropped from $N_{max}=10,000$ to $N_{opt}=3,430$.}
    \label{fig:histogram_cifar10_10000B_after_remove_empty_bins}
\end{figure}

\textbf{Population Initialization.}
The starting point of the DE algorithm is population initialization, and here, we designed a heuristic blocked and compressed population initialization based on Histogram-based Block (HBB) on pre-trained parameters.
In fact, the distance between parameters is very close enough to tie the weights.
Therefore, we can divide the weights into bins using the histogram algorithm. 
Let us assume the weights in DNNs are $W=[w_1, w_2, ..., w_D]$ where $D$ is the total number of trainable parameters in the model. By a given $N_{bins}$, parameters are placed into $N_{bins}$ bins, and, after removing empty bins, the indexes of parameters placed in $bin_{i}$ are stored in into $N_{blocks}$ blocks. The proposed approach divides parameters into equal sizes of $N_{bins}$ where the contents of each bin are blocked, and an average of dimensions is calculated and replaced into the blocked vector of parameters. Fig. \ref{fig:histogram_cifar10_10000B_before_remove_empty_bins} demonstrates a histogram of ResNet-18 model parameters trained on CIFAR-10 in the range of its min and max values for $N_{bins} = 10,000$. 
The sparsity between the histogram bars shows the bins are empty, meaning there is no association with any parameters to place within its boundaries. After removing empty bins, the number of bins reduces to final $N_{opt} = 3,430$ where Fig. \ref{fig:histogram_cifar10_10000B_after_remove_empty_bins} shows the final representation.
In the end, the size of the blocked population is $[NP, N_{blocks}]$. A candidate solution in population includes $Wb=[w_1, w_2, ..., w_{D^{'}}]$ where $D^{'} = N_{blocks}$ for simplicity.

\textbf{Unblocking.} Since we defined the blocks of parameters using HBB, the unblocking is a transformation problem to transfer the blocked parameters to the original space by hardly copying into the parameters of the DNN model. As previously discussed in \cite{lecun1989generalization}, weight sharing is a technique to generalize the learned knowledge into the neural network, which increases the performance. Once the tied parameters are defined by the blocks, the transformation from a single parameter in the candidate solution is a hard-wired process to the weights and biases located in the same block. In other words, hard weight sharing is the name given to this kind of weight tying, in which the weights are shared among the layers of a neural network, increasing efficiency in parameter storage and utilization.

\textbf{Block DE.} After initializing the blocked population, the model is passed to the gradient-free optimizer DE to fine-tune the blocked parameters for a given number of function evaluations $NFE$. The objective function is to maximize the performance of the DNN model in terms of the F1-score metric, using the optimization process as summarized in Algorithm \ref{pseudoHBDE}.

\begin{algorithm}
\caption{Deep Neural Network Training by Histogram-based Block DE.}
\label{pseudoHBDE}
\hspace*{\algorithmicindent} \textbf{Input:} expected dimension $D_{exp}$, a deep neural network $w$, maximum functions evaluations $Max_{NFE}$, number of population $NP$, training dataset $DS$ \\
\hspace*{\algorithmicindent} \textbf{Output:} A fine-tuned network $w$

\begin{algorithmic}[1]
\State Fine-tune model $w$ on dataset $DS$.
\State let $\rho_{w} = [W^{1}, b^{1}, ..., W^{l},b^{l}]$ 
\State let $N_{bin} = D_{exp}$
\State $bins$ $\leftarrow$ histogram($N_{bin}$, $\rho_{w}$)
\State Remove empty $bins$.
\State $D^{'} \leftarrow size(bins)$
\State Initialize $NP$ population $\overrightarrow{P}$ with uniform random values in the boundaries of each $bins$ in dimension of $D^{'}$.
\State $blocks$ $\leftarrow$ Find locations (indices) of each $\rho_{w}$ in $bins$
\While{Termination condition not reached ($NFE<Max_{NFE}$}
\For{$i \gets 1 \ \textbf{to} \ NP$}
    \State $\overrightarrow{X}_i \leftarrow$ update $\overrightarrow{P}_{i}$ by mutation and crossover ops.
    \State $\overrightarrow{uX}_i \leftarrow unblocker(X_{i})$
    \State Evaluate the fitness F1-score({$w, f(\overrightarrow{uX}_i)$)}
    \If{$f(\overrightarrow{uX}_i)>f(\overrightarrow{P}_i)$} \Comment{Maximization}
       \State $\overrightarrow{P}_i \gets \overrightarrow{uX}_i$
    \EndIf
\EndFor
\State $NFE \leftarrow NFE + NP$
\EndWhile
\end{algorithmic}
\end{algorithm}

\section{Experimental analysis}\label{sec:experiments}

Our experiments study the performance of HBDE in optimizing the ResNet-18 model on two well-known benchmark datasets. The ResNet is a deep learning architecture that was introduced by He \textit{et al.} \cite{He_2016_CVPR}. 
The total number of trainable parameters in ResNet-18 is $\sim$11M without a fully connected network. 

\subsection{CIFAR-10/100}
\begin{figure}[t]
    \centering
    \includegraphics[scale=0.45]{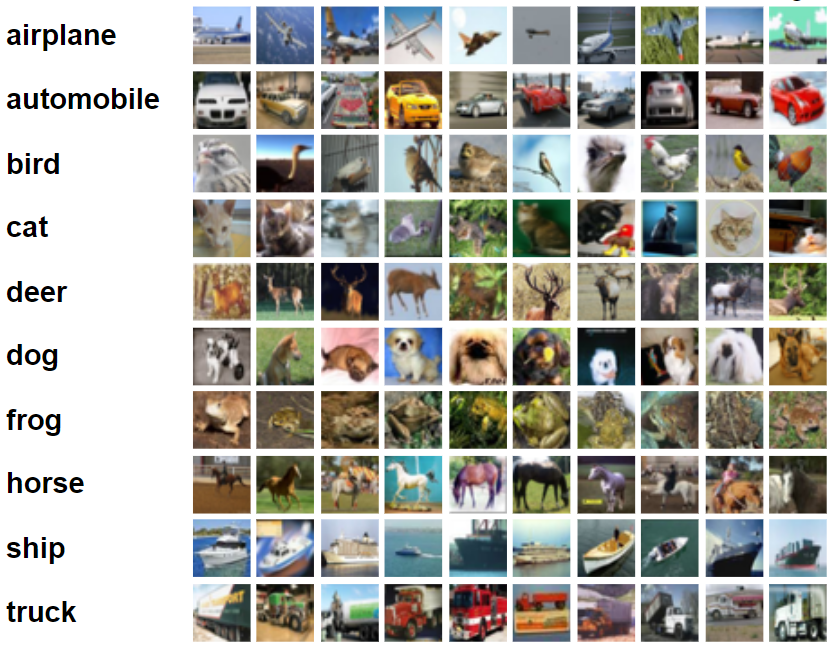}
    \caption{Sample images from CIFAR-10 dataset.}
    \label{fig:cifar10_samples}
\end{figure}

The CIFAR-10 and CIFAR-100 are image classification benchmarks that have 10 and 100 distinct classes, respectively. They include 50K $32 \times 32$ color images for train and 10K images for test \cite{krizhevsky2009learning}.
A few samples from CIFAR-10 are shown in Fig. \ref{fig:cifar10_samples}. 
The ResNet-18 model $w$ is pre-trained on the ImageNet 2012 dataset \cite{russakovsky2015imagenet}. Then, the ResNet-18 model $w$ is fine-tuned, evaluated on circular super-batches of 10K training images, and finally evaluated on test images, reported by the F1-score metric. 

The training process includes the following steps:
\begin{enumerate}
    \item Loading the pre-trained model $w$, which is trained by the ImageNet dataset. \cite{deng2009imagenet}.
    \item Fine-tuning model $w$ by a gradient-based optimizer, to get $w^{'}$. We use Adam with a learning rate of $lr=10^{-3}$, a weight decay of zero, and a cross-entropy loss function formulated as follows:
    \begin{equation}
        \mathcal{L}_{cross-entropy}(\hat{y},y)=-\sum_{k}^K y_{i}^{(k)}\log(\hat{y}^{(k)}).
    \end{equation}
    Where $K$ is the number of classes in the dataset.
        \begin{table}[!h]
        \caption{DE and HBDE parameters setting.}
        \label{tab:DE_parameter_setting}
        \centering
        \begin{tabular}{l|ll}
            \hline
            Parameters & Classic DE  &HBDE\\ \hline
            $F$ & 0.5  &0.5  
\\ & & \\
            $Cr$ & 0.9  &0.9  
\\
            Strategy & rand/1/bin  &rand/1/bin  \\ \hline
 $D_{exp}$ ($N_{bins}$)& -& $10,000$\\\hline
        \end{tabular}
    \end{table} 
    
    \item Optimizing the model by the proposed HBDE and get $w^{'''}$. The parameters setting is reviewed in Table \ref{tab:DE_parameter_setting} where $F$ is the mutation rate, and $Cr$ is crossover probability.

\end{enumerate}

\begin{figure}[h]
    \centering
    \begin{subfigure}[b]{\columnwidth}
        \centering
        \includegraphics[scale=0.45]{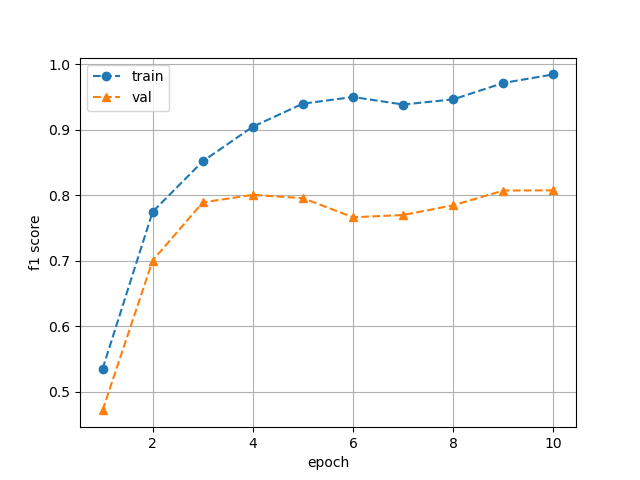}
        \caption{CIFAR-10.}
    \end{subfigure}%
    \\
    \begin{subfigure}[b]{\columnwidth}
        \centering
        \includegraphics[scale=0.45]{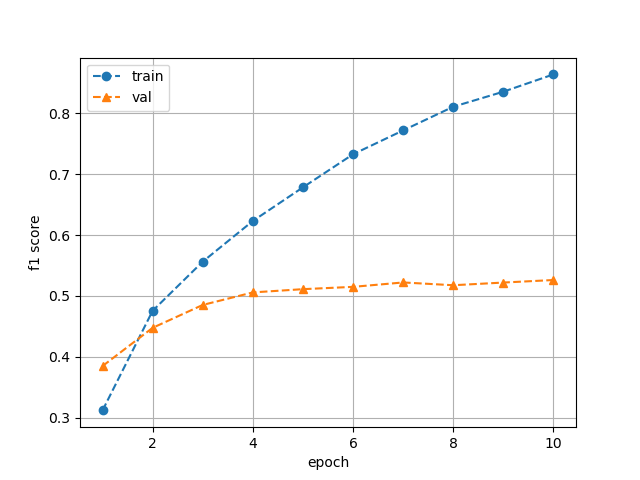}
        \caption{CIFAR-100.}
    \end{subfigure}
    \caption{Gradient-based pre-training.}
\end{figure}

A comparison is made between the places where the model is pre-trained with Adam and fine-tuned with classic DE and HBDE with the parameter settings in Table \ref{tab:DE_parameter_setting}. For a fair competition, each gradient-free optimization algorithm is terminated by reaching $Max_{NFE} = 100,000$ number of objective function evaluations.

\begin{table}[h]
    \centering
    \caption{Results of training ResNet-18 model evaluated on CIFAR-10 based on the F1-score metric. The size of parameters in the search space is indicated as $D$.}
\begin{tabular}{@{}c|cccc@{}}
\toprule
Algorithm  & \#Params ($D$) & $f(x)$         & Train F1-score & Test F1-score  \\ \midrule
Adam       & 11M           & -              & 97.23          & 80.76          \\ \midrule
Classic DE & 11M           & 99.24          & 98.31          & 81.34          \\
HBDE       & \textbf{3.4K} & \textbf{99.44} & \textbf{98.55} & \textbf{81.83} \\ \bottomrule
\end{tabular}
    
    \label{tab:cifar10}
\end{table}

Table \ref{tab:cifar10} presents the findings of the CIFAR-10 dataset. The pre-trained model by Adam results in $80.76\%$ test F1-score. In the next step, DE and the proposed HBDE fine-tune the model based on maximizing the F1-score. The classic DE resulted in $81.34\%$ with an improvement of $0.82\%$ in the performance of the pre-trained model. On the other hand, our proposed HBDE could reach $81.83\%$ F1-score, which is $1.07\%$ better performance compared to the pre-trained model and ultimately $0.5\%$ better than its parent, classic DE algorithm.

\begin{table}[h]
    \centering
    \caption{Results of training ResNet-18 model evaluated on CIFAR-100 based on the F1-score metric. The size of parameters in the search space is indicated as $D$.}

\begin{tabular}{@{}c|cccc@{}}
\toprule
Algorithm  & \#Params ($D$) & $f(x)$         & Train F1-score & Test F1-score  \\ \midrule
Adam       & $11$M   &-& 90.76 & 52.61 \\ \midrule
Classic DE & $11$M   &92.67& 91.35& 52.69\\
HBDE       & \textbf{3.5K}&\textbf{93.50}& \textbf{91.60}& \textbf{53.30} \\ \bottomrule
\end{tabular}
    
    \label{tab:cifar100}
\end{table}

Table \ref{tab:cifar100} displays the findings from the CIFAR-100 dataset. In more detail, the model is pre-trained by Adam and resulted in $52.61\%$ and stopped by reaching the maximum number of epochs; then, in a separate optimization, DE optimized and is evaluated as $52.69\%$ F1-score a similar performance to F1-score, whereas HBDE is much more successful with $53.30\%$ in terms of F1-score metric. 

From Tables \ref{tab:cifar10} and \ref{tab:cifar100}, one can realize that the proposed HBDE has a dramatic convergence rate in the initial steps as the reduced and blocked population is randomly initiated within the blocks, showcasing the effectiveness of block utilization for gradient-free optimization algorithms. It is worth mentioning that the proposed method is explored in a 3K times smaller space, and tied weights located in blocks reconstruct the model with a single parameter.

\begin{figure}[h!]
    \centering
    \begin{subfigure}[h]{\columnwidth}
        \centering
        \includegraphics[scale=0.5]{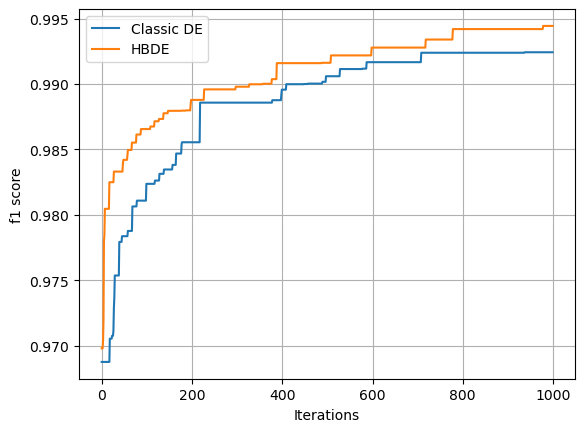}
        \caption{CIFAR-10.}
    \end{subfigure}%
    \\
    \begin{subfigure}[h]{\columnwidth}
        \centering
        \includegraphics[scale=0.5]{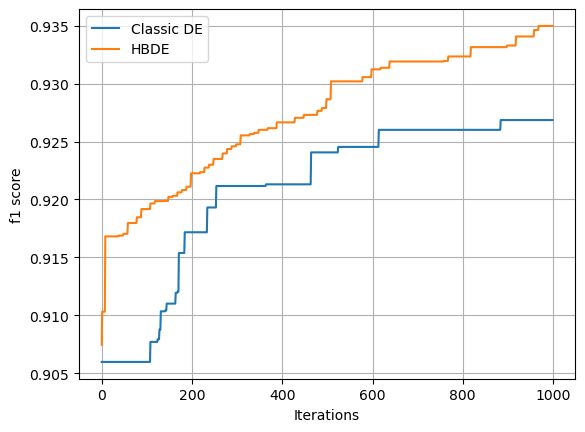}
        \caption{CIFAR-100.}
    \end{subfigure}
    \caption{Convergence plots for meta-heuristics on optimizing model evaluated on circular super-batches of data.}
\end{figure}

\section{Conclusion Remarks}\label{sec:conclusion}

This paper presents a novel massive dimension reduction approach to tackle the sparsity in deep neural networks during gradient-free optimization. Our method blocks large numbers of parameters and reduces the search space to thousands of times smaller, allowing for straightforward utilization of meta-heuristics to train deep neural networks.

Optimizing parameters in the ResNet-18 model with 11M trainable parameters by an evolutionary algorithm is inefficient regarding resources and ineffective in a huge-scale search space. We demonstrate that the accuracy does not drop by blocking weights in a deep neural network. 

Using our proposed blocking approach, training, fine-tuning, and storing deep neural networks is much more efficient during inference time. Similar to other weight-sharing techniques in the literature, this approach stays with the same architecture and hardly shares the weights across the whole network.

A major advantage of our approach is that huge dimension reduction enables meta-heuristic algorithms to take a step into optimizing huge-scale neural networks for different tasks without the limitations of traditional differentiable loss functions. In other words, it guarantees the applicability of gradient-free optimization, which has many benefits over traditional gradient-based optimization, such as trapping into local optima and stagnation.
In addition, it opens the door for multi-objective training, with the benefit of multi-modal, multi-task, and multi-loss training.
The experiments prove that the proposed HBDE outperforms Adam and its parent, classic DE, algorithm, which is evaluated on CIFAR-10 and CIFAR-100 datasets. 

In future works, we aim to find the optimized number of bins $N_{bin}$ and larger blocks to reduce the search space dimension further to result in a tiny trainable DNN. Given the successful approval of the block concept in the hybridization of gradient-based and gradient-free optimization, we are ambitious to eliminate reliance on pre-trained parameters by gradient-based optimization and train any deep neural network architecture from scratch only with gradient-free optimization in an efficient strategy. Moreover, we are able to break weighted sum (average) single-objective functions to multi-objective functions, for instance, in our work, convert \textit{F1-score} (see Eq. \ref{eq:f1score}) to \textit{Precision} (see Eq. \ref{eq:PR}) and \textit{Recall} (see Eq. \ref{eq:RE}) and optimize by any multi-objective gradient-free optimization algorithms such as NSGA-II \cite{deb2002fast}, efficiently and effectively.

\bibliographystyle{ieeetr}
\bibliography{references.bib}

\begin{thebibliography}{10}

\bibitem{kingma2014adam}
D.~P. Kingma and J.~Ba, ``Adam: A method for stochastic optimization,'' {\em arXiv preprint arXiv:1412.6980}, 2014.

\bibitem{gupta2023explainable}
L.~K. Gupta, D.~Koundal, and S.~Mongia, ``Explainable methods for image-based deep learning: a review,'' {\em Archives of Computational Methods in Engineering}, vol.~30, no.~4, pp.~2651--2666, 2023.

\bibitem{kaveh2023application}
M.~Kaveh and M.~S. Mesgari, ``Application of meta-heuristic algorithms for training neural networks and deep learning architectures: A comprehensive review,'' {\em Neural Processing Letters}, vol.~55, no.~4, pp.~4519--4622, 2023.

\bibitem{al2023boosting}
M.~A. Al-Betar, M.~A. Awadallah, I.~A. Doush, O.~A. Alomari, A.~K. Abasi, S.~N. Makhadmeh, and Z.~A.~A. Alyasseri, ``Boosting the training of neural networks through hybrid metaheuristics,'' {\em Cluster Computing}, vol.~26, no.~3, pp.~1821--1843, 2023.

\bibitem{ansari2020hybrid}
A.~Ansari, I.~S. Ahmad, A.~A. Bakar, and M.~R. Yaakub, ``A hybrid metaheuristic method in training artificial neural network for bankruptcy prediction,'' {\em IEEE access}, vol.~8, pp.~176640--176650, 2020.

\bibitem{HybridPSOCNN}
Y.~Chhabra, S.~Varshney, and A.~Wadhwa, ``Hybrid particle swarm training for convolution neural network {(CNN)},'' in {\em 2017 Tenth International Conference on Contemporary Computing (IC3)}, pp.~1--3, 2017.

\bibitem{ganjefar2017training}
S.~Ganjefar and M.~Tofighi, ``Training qubit neural network with hybrid genetic algorithm and gradient descent for indirect adaptive controller design,'' {\em Engineering Applications of Artificial Intelligence}, vol.~65, pp.~346--360, 2017.

\bibitem{yaghini2013hybrid}
M.~Yaghini, M.~M. Khoshraftar, and M.~Fallahi, ``A hybrid algorithm for artificial neural network training,'' {\em Engineering Applications of Artificial Intelligence}, vol.~26, no.~1, pp.~293--301, 2013.

\bibitem{rere2016metaheuristic}
L.~Rere, M.~I. Fanany, A.~M. Arymurthy, {\em et~al.}, ``Metaheuristic algorithms for convolution neural network,'' {\em Computational intelligence and neuroscience}, vol.~2016, 2016.

\bibitem{akay2022comprehensive}
B.~Akay, D.~Karaboga, and R.~Akay, ``A comprehensive survey on optimizing deep learning models by metaheuristics,'' {\em Artificial Intelligence Review}, pp.~1--66, 2022.

\bibitem{Rokhsatyazdi2023ANNCS}
E.~Rokhsatyazdi, S.~Rahnamayan, S.~Z. Miyandoab, A.~A. Bidgoli, and H.~Tizhoosh, ``Training artificial neural networks by coordinate search algorithm,'' in {\em 2023 IEEE Symposium Series on Computational Intelligence (SSCI)}, pp.~1540--1546, 2023.

\bibitem{He_2016_CVPR}
K.~He, X.~Zhang, S.~Ren, and J.~Sun, ``Deep residual learning for image recognition,'' in {\em Proceedings of the IEEE Conference on Computer Vision and Pattern Recognition (CVPR)}, June 2016.

\bibitem{radford2019language}
A.~Radford, J.~Wu, R.~Child, D.~Luan, D.~Amodei, and I.~Sutskever, ``Language models are unsupervised multitask learners,'' 2019.

\bibitem{rumelhart1986general}
D.~E. Rumelhart, G.~E. Hinton, J.~L. McClelland, {\em et~al.}, ``A general framework for parallel distributed processing,'' {\em Parallel distributed processing: Explorations in the microstructure of cognition}, vol.~1, no.~45-76, p.~26, 1986.

\bibitem{lecun1989generalization}
Y.~LeCun {\em et~al.}, ``Generalization and network design strategies,'' {\em Connectionism in perspective}, vol.~19, no.~143-155, p.~18, 1989.

\bibitem{nowlan2018simplifying}
S.~J. Nowlan and G.~E. Hinton, ``Simplifying neural networks by soft weight sharing,'' in {\em The Mathematics of Generalization}, pp.~373--394, CRC Press, 2018.

\bibitem{DE1997}
R.~Storn and K.~Price, ``Differential evolution – a simple and efficient heuristic for global optimization over continuous spaces,'' {\em J. of Global Optimization}, vol.~11, p.~341–359, Dec. 1997.

\bibitem{KhosrowshahliBlockDE2023}
R.~Khosrowshahli and S.~Rahnamayan, ``Block differential evolution,'' in {\em 2023 IEEE Congress on Evolutionary Computation (CEC)}, pp.~1--8, 2023.

\bibitem{9283201}
D.~Z. Farsa and S.~Rahnamayan, ``Discrete coordinate descent {(DCD)},'' in {\em 2020 IEEE International Conference on Systems, Man, and Cybernetics (SMC)}, pp.~184--190, 2020.

\bibitem{bodner2021gradfreebits}
B.~J. Bodner, G.~B. Shalom, and E.~Treister, ``Gradfreebits: Gradient free bit allocation for dynamic low precision neural networks,'' {\em arXiv preprint arXiv:2102.09298}, 2021.

\bibitem{ng2005projective}
E.~K.~K. Ng, A.-C. Fu, and R.-W. Wong, ``Projective clustering by histograms,'' {\em IEEE Transactions on Knowledge and Data Engineering}, vol.~17, no.~3, pp.~369--383, 2005.

\bibitem{krizhevsky2009learning}
A.~Krizhevsky, G.~Hinton, {\em et~al.}, ``Learning multiple layers of features from tiny images,'' 2009.

\bibitem{russakovsky2015imagenet}
O.~Russakovsky, J.~Deng, H.~Su, J.~Krause, S.~Satheesh, S.~Ma, Z.~Huang, A.~Karpathy, A.~Khosla, M.~Bernstein, {\em et~al.}, ``Imagenet large scale visual recognition challenge,'' {\em International journal of computer vision}, vol.~115, pp.~211--252, 2015.

\bibitem{deng2009imagenet}
J.~Deng, W.~Dong, R.~Socher, L.-J. Li, K.~Li, and L.~Fei-Fei, ``Imagenet: A large-scale hierarchical image database,'' in {\em 2009 IEEE conference on computer vision and pattern recognition}, pp.~248--255, Ieee, 2009.

\bibitem{deb2002fast}
K.~Deb, A.~Pratap, S.~Agarwal, and T.~Meyarivan, ``A fast and elitist multiobjective genetic algorithm: Nsga-ii,'' {\em IEEE transactions on evolutionary computation}, vol.~6, no.~2, pp.~182--197, 2002.

\end{thebibliography}

\end{document}